\def\year{2016}\relax
\documentclass[letterpaper]{article}
\usepackage{aaai16}
\usepackage{times}
\usepackage{helvet}
\usepackage{courier}
\usepackage{amssymb}
\usepackage{amsmath}
\usepackage{amsfonts}
\usepackage{graphicx}
\usepackage{url}

\begin{document}
%
\title{A Deep Architecture for Semantic Matching \\with Multiple Positional Sentence Representations}
\author{Shengxian Wan$^{*}$, Yanyan Lan$^\dag$, Jiafeng Guo$^\dag$, Jun Xu$^\dag$, Liang Pang$^{*}$, \and Xueqi Cheng$^\dag$\\
CAS Key Lab of Network Data Science and Technology\\
Institute of Computing Technology, Chinese Academy of Sciences, China\\
$^*$\{wanshengxian, pangliang\}@software.ict.ac.cn, $^\dag$\{lanyanyan, guojiafeng, junxu, cxq\}@ict.ac.cn \\
}
\maketitle
\begin{abstract}
\begin{quote}
Matching natural language sentences is central for many applications such as information retrieval and question answering.
Existing deep models rely on a single sentence representation or multiple granularity representations for matching. However, such methods cannot well capture the contextualized local information in the matching process. To tackle this problem, we present a new deep architecture to match two sentences with multiple positional sentence representations. Specifically, each positional sentence representation is a sentence representation at this position, generated by a bidirectional long short term memory (Bi-LSTM). The matching score is finally produced by aggregating interactions between these different positional sentence representations, through $k$-Max pooling and a multi-layer perceptron. Our model has several advantages: (1) By using Bi-LSTM, rich context of the whole sentence is leveraged to capture the contextualized local information in each positional sentence representation; (2) By matching with multiple positional sentence representations, it is flexible to aggregate different important contextualized local information in a sentence to support the matching; (3) Experiments on different tasks such as  question answering and sentence completion demonstrate the superiority of our model.
\end{quote}
\end{abstract}

\section{Introduction}
Semantic matching is a critical task for many applications in natural language processing (NLP), such as information retrieval \cite{INR-035}, question answering \cite{Berger:2000:BLC:345508.345576} and paraphrase identification \cite{dolan2004unsupervised}.
Taking question answering as an example, given a pair of question and answer, a matching function is required to determine the matching degree between these two sentences.

Recently, deep neural network based models have been applied in this area and achieved some important progresses.
A lot of deep models follow the paradigm to first represent the whole sentence to a single distributed representation, and then compute similarities between the two vectors to output the matching score. Examples include DSSM \cite{huang2013learning}, CDSMM \cite{export:226585}, ARC-I \cite{DBLP:conf/nips/HuLLC14}, CNTN \cite{DBLP:conf/ijcai/QiuH15} and LSTM-RNN \cite{palangi2015deep}.
In general, this paradigm is quite straightforward and easy to implement, however, the main disadvantage lies in that important local information is lost when compressing such a complicated sentence into a single vector. Taking a question and two answers as an example:

{\em Q: `` Which teams won top three in the World Cup?}"

{\em A1: `` Germany is the champion of the World Cup.}"

{\em A2: ``The top three of the European Cup are Spain, Netherlands and Germany. }"

We can see that the keywords such as {\em``top three''} and {\em``World Cup''} are very important to determine which answer (between {\em A1} and {\em A2}) is better for {\em Q}. When attending to {\em``top three''}, obviously {\em A2} is better than  {\em A1}; while if attending to {\em``World Cup''}, we can get an opposite conclusion.
However, single sentence representation methods cannot well capture such important local information, by directly representing a complicated sentence as a single compact vector \cite{bahdanau2014neural}.

Some other works focus on taking multiple granularity, e.g.~word, phrase, and sentence level representations, into consideration for the matching process. Examples include ARC-II~\cite{DBLP:conf/nips/HuLLC14}, RAE \cite{socher2011dynamic}, DeepMatch~\cite{DBLP:conf/nips/LuL13}, Bi-CNN-MI~\cite{DBLP:conf/naacl/YinS15} and MultiGranCNN~\cite{DBLP:conf/acl/YinS15}. They can alleviate the above problem, but are still far from completely solving the matching problem.  
That is because they are limited to well capture the contextualized local information, by directly involving word and phrase level representations. Taking the following answer as an example:

{\em A3: ``The top three attendees of the European Cup are from Germany, France and Spain.}"

Obviously, {\em A2} is better than {\em A3} with respect to $Q$, although both of them have the important keywords {\em``top three''}. This is because the two terms of {\em``top three"} have different meanings from the whole sentence perspective. {\em``top three"} in {\em A2} focuses on talking about top three football teams, while that in {\em A3} is indicating the top three attendees from different countries. However, existing multiple granularity deep models cannot well distinguish the two {\em``top three''}s. This is mainly because the word/phrase level representations are local (usually depend on contexts in a fixed window size), thus limited to reflect the true meanings of these words/phrases (e.g.~{\em top three}) from the perspective of the whole sentence.

From the above analysis, we can see that the matching degree between two sentences requires sentence representations from contextualized local perspectives. This key observation motivates us to conduct matching from multiple views of a sentence. That is to say, we can use multiple sentence representations in the matching process, with each sentence representation focusing on different local information.

In this paper, we propose a new deep neural network architecture for semantic matching with multiple positional sentence representations, namely MV-LSTM.
Firstly, each positional sentence representation is defined as a sentence representation at one position. We adopt a bidirectional long short term memory (Bi-LSTM) to generate such positional sentence representations in this paper. Specifically for each position, Bi-LSTM can obtain two hidden vectors to reflect the meaning of the whole sentence from two directions when attending to this position. The positional sentence representation can be generated by concatenating them directly.
The second step is to model the interactions between those positional sentence representations. In this paper, three different operations are adopted to model the interactions: cosine, bilinear, and tensor layer. Finally, we adopt a $k$-Max pooling strategy to automatically select the top $k$ strongest interaction signals, and aggregate them to produce the final matching score by a multi-layer perceptron (MLP).
Our model is end to end, and all the parameters are learned automatically from the training data, by BackPropagation and Stochastic Gradient Descent.

We can see that our model can well capture contextualized local information in the matching process. Compared with single sentence representation methods, MV-LSTM can well capture important local information by introducing multiple positional sentence representations. While compared with multiple granularity deep models, MV-LSTM has leveraged rich context to determine the importance of the local information by using Bi-LSTM to generate each positional sentence representation.
Finally, we conduct extensive experiments on two tasks, i.e.~question answering and sentence completion, to validate these arguments. Our experimental results show that MV-LSTM can outperform several existing baselines on both tasks, including ARC-I , ARC-II, CNTN, DeepMatch, RAE, MultiGranCNN and LSTM-RNN.

The contribution of this work lies in three folds:
\begin{itemize}
\item the proposal of matching with multiple positional sentence representations, to capture important contextualized local information;
\item a new deep architecture to aggregate the interactions of those positional sentence representations for semantic matching, with each positional sentence representation generated by a Bi-LSTM;
\item experiments on two tasks (i.e.~question answering and sentence completion) to show the benefits of our model.
\end{itemize}

\section{Our Approach}
\begin{figure}[t]
\centering
\includegraphics[width=0.47\textwidth]{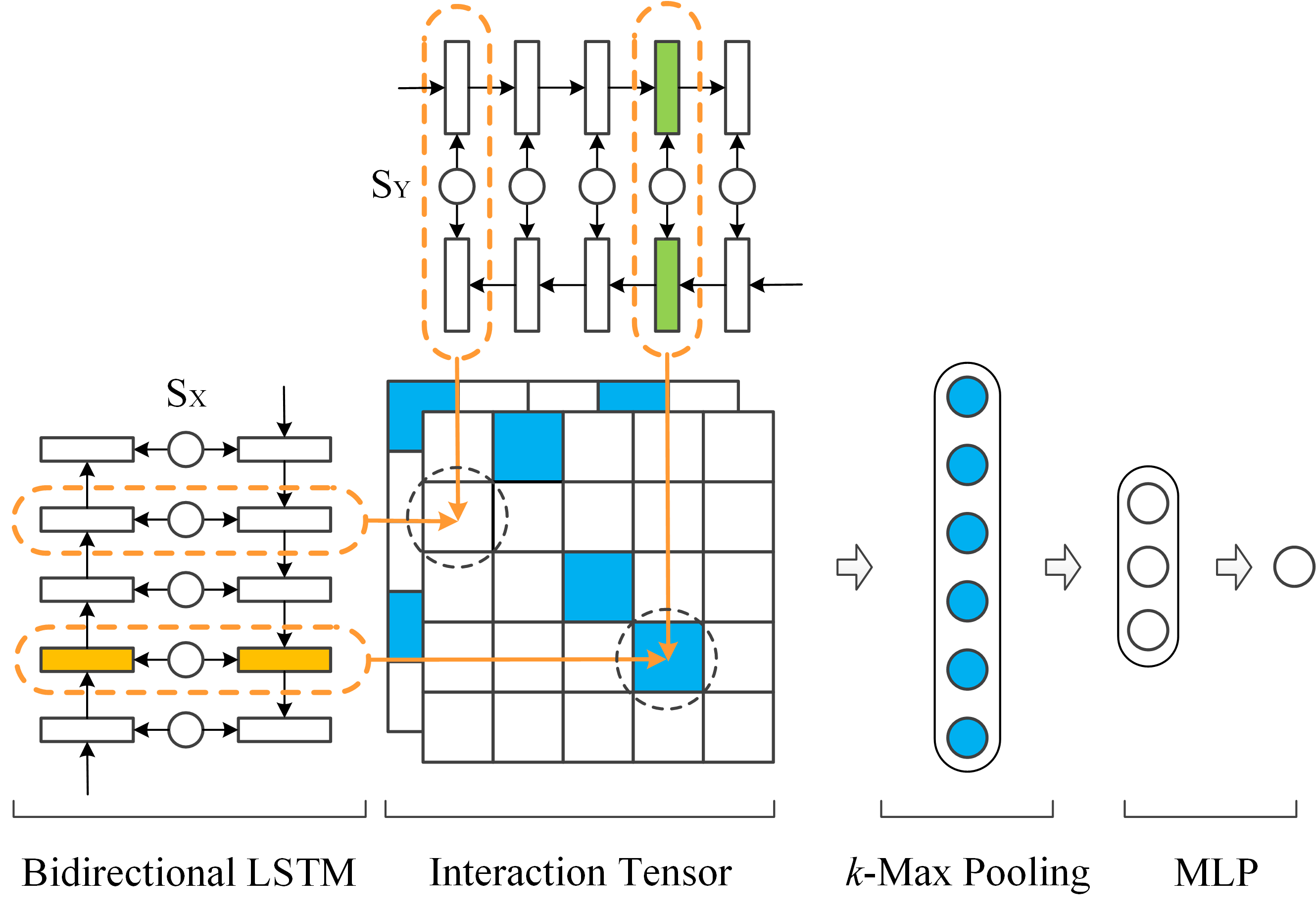}
\caption{Illustration of MV-LSTM. $S_X$ and $S_Y$ are the input sentences. Positional sentence representations (denoted as the dashed orange box) are first obtained by a Bi-LSTM. $k$-Max pooling then selects the top $k$ interactions from each interaction matrix (denoted as the blue grids in the graph). The matching score is finally computed through a multi-layer perceptron.}
\label{fig:model}
\end{figure}

In this section, we present our new deep architecture for matching two sentences with multiple positional sentence representations, namely MV-LSTM.
As illustrated in Figure \ref{fig:model}, MV-LSTM consists of three parts: Firstly, each positional sentence representation is a sentence representation at one position, generated by a bidirectional long short term memory (Bi-LSTM); Secondly, the interactions between different positional sentence representations form a similarity matrix/tensor by different similarity functions; Lastly, the final matching score is produced by aggregating such interactions through $k$-Max pooling and a multilayer perceptron.
\subsection{Step 1: Positional Sentence Representation}
Each positional sentence representation requires to reflect the representation of the whole sentence when attending to this position.
Therefore, it is natural to use a Bi-LSTM to generate such representation because LSTM can both capture long and short term dependencies in the sentences. Besides, it has a nice property to emphasize nearby words in the representation process \cite{bahdanau2014neural}.

Firstly, we give an introduction to LSTM and Bi-LSTM.
Long short term memory (LSTM) is an advanced type of Recurrent Neural Network by further using memory cells and gates to learn long term dependencies within a sequence \cite{Hochreiter:1997:LSM:1246443.1246450}. LSTM has several variants \cite{2015arXiv150304069G}, and we adopt one common implementation used in \cite{graves2013speech}, but without peephole connections, as did in \cite{palangi2015deep}. Given an input sentence $S=(x_0,x_1,\cdots,x_T)$, where $x_t$ is the word embedding at position $t$. LSTM outputs a representation $h_t$ for position $t$ as follows.
\begin{eqnarray*}
 &&i_t=\sigma(W_{xi}x_t+W_{hi}h_{t-1}+b_{i}),\\
 &&f_t=\sigma(W_{xf}x_t+W_{hf}h_{t-1}+b_{f}),\\
  &&c_t=f_tc_{t-1}+i_t\tanh(W_{xc}x_t+W_{hc}h_{t-1}+b_{c}),\\
   &&o_t=\sigma(W_{xo}x_t+W_{ho}h_{t-1}+b_{o}),\\
   &&h_t=o_t\tanh(c_t)
\end{eqnarray*}
where $i, f, o$ denote the input, forget and output gates respectively. $c$ is the information stored in memory cells and $h$ is the representation.
Compared with single directional LSTM, bidirectional LSTM utilizes both the previous and future context, by processing the data from two directions with two separate LSTMs \cite{Schuster:1997:BRN:2198065.2205129}. One LSTM processes the input sequence in the forward direction while the other processes the input in the reverse direction. Therefore, we can obtain two vectors $\overrightarrow{h_t}$ and $\overleftarrow{h_t}$ for each position.

Intuitively, $\overrightarrow{h_t}$ and $\overleftarrow{h_t}$ reflect the meaning of the whole sentence from two directions when attending to this position, therefore it is reasonable to define the positional sentence representation as the combination of them. Specifically, for each position $t$, the $t$-th positional sentence representation $p_t$ is generated by concatenating $\overrightarrow{h_t}$ and $\overleftarrow{h_{t}}$, i.e.~$p_t=[\overrightarrow{h_t}^T,\overleftarrow{h_{t}}^T]^T$, where $(\cdot)^T$ stands for the transposition operation which will also be used later.
\subsection{Step 2: Interactions Between Two Sentences}
On the basis of positional sentence representations, we can model the interactions between a pair of sentences from different positions. Many kinds of similarity functions can be used for modeling the interactions between $p_{Xi}$  and $p_{Yj}$, where $p_{Xi}$ and $p_{Yj}$ stand for the $i$ and $j$-th positional sentence representations for two sentences $S_X$ and $S_Y$, respectively. In this paper, we use three similarity functions, including cosine, bilinear and tensor layer. Given two vectors $u$ and $v$, the three functions will output the similarity score $s(u,v)$ as follows.

{\bf Cosine} is a common function to model interactions. The similarity score is viewed as the angle of two vectors:
\begin{equation*}
s(u,v)=\frac{u^Tv}{||u||\cdot||v||},
\end{equation*}
where $||\cdot||$ stands for the L2 norm.

{\bf Bilinear} further considers interactions between different dimensions, thus can capture more complicated interactions as compared with cosine. Specifically, the similarity score is computed as follows:
\begin{equation*}
s(u,v)={u^T}Mv+b,
\end{equation*}
where $M$ is the matrix to reweight the interactions between different dimensions, and $b$ is the bias. When applying bilinear to compute the interaction between two corresponding positional sentence representations $p_{Xi}$ and $p_{Yj}$ for sentence $S_X$ and $S_Y$, obviously bilinear can well capture the interleaving interactions between $\overrightarrow{h_{Xi}}$ and $\overleftarrow{h_{Yj}}$, while cosine cannot. Therefore bilinear can capture more meaningful interactions between two positional sentence representations, compared with cosine.

{\bf Tensor Layer} is more powerful than the above two functions, which can roll back to other similarity metrics such as bilinear and dot product. It has also shown great superiority in modeling interactions between two vectors \cite{socher2013recursive,socher2013reasoning,DBLP:conf/ijcai/QiuH15}. That's why we choose it as an interaction function in this paper.
Other than outputing a scalar value as bilinear and cosine do, tensor layer outputs a vector, as described as follows.
\begin{equation*}
s(u,v)= f(u^TM^{[1:c]}v+W_{uv}\begin{bmatrix}
		u\\
		v
	\end{bmatrix}+b),
\end{equation*}
where $M^i, i\in[1,...,c]$ is one slice of the tensor parameters, $W_{uv}$ and $b$ are parameters of the linear part. $f$ is a non-linear function, and we use rectifier $f(z)=\max(0,z)$~\cite{glorot2011deep} in this paper, since it always outputs a positive value which is compatible as a similarity.

We can see that the outputs of the former two similarity functions (i.e. cosine and bilinear) are both interaction matrices,
while the tensor layer will output an interaction tensor, as illustrated in Figure 1.

\subsection{Step 3: Interaction Aggregation}

Now we introduce the third step of our architecture, i.e.~how to integrate such interactions between different positional sentence representations to output a matching score for two sentences.

\subsubsection{$k$-Max Pooling}
The matching between two sentences is usually determined by some strong interaction signals. Therefore, we use $k$-Max pooling to automatically extract top $k$ strongest interactions in the matrix/tensor, similar to \cite{KalchbrennerACL2014}. Specifically for the interaction matrix, we scan the whole matrix and the top $k$ values are directly returned to form a vector $q$ according to the descending order. While for the interaction tensor, the top $k$ values of each slice of the tensor are returned to form a vector. Finally, these vectors are further concatenated to a single vector $q$.

$k$-Max pooling is meaningful: suppose we use cosine similarity, when $k=1$, it directly outputs the largest interaction, which means that only the ``best matching position" is considered in our model; while $k$ is larger than 1 means that we utilize the top $k$ matching positions to conduct semantic matching. Therefore, it is easy to detect where the best matching position lies, and whether we need to aggregate multiple interactions from different positions for matching. Our experiments show that the best matching position is usually not the first or last one, and better results can be obtained by leveraging matchings on multiple positions.

\subsubsection{MultiLayer Perception}

Finally, we use a MLP to output the matching score by aggregating such strong interaction signals filtered by $k$-Max pooling. Specifically, the feature vector $q$ obtained by $k$-Max pooling is first feed into a full connection hidden layer to obtain a higher level representation $r$. Then the matching score $s$ is obtained by a linear transformation:
\begin{equation*}
r= f(W_rq+b_r),\,\,s=W_sr+b_s,
\end{equation*}
where $W_r$ and $W_s$ stands for the parameter matrices, and $b_r$ and $b_s$ are corresponding biases.

\subsection{Model Training}

For different tasks, we need to utilize different loss functions to train our model.
For example, if the task is formalized as a ranking problem, we can utilize pairwise ranking loss such as hinge loss for training. Given a triple $(S_X,S_Y^+,S_Y^-)$, where $S_Y^+$ is ranked higher than $S_Y^-$ when matching with $S_X$, the loss function is defined as:
\begin{equation*}
\mathcal{L}(S_X, S^+_Y, S^-_Y) = \max(0, 1-s(S_X, S_Y^+)+s(S_X, S_Y^-))
\end{equation*}
where $s(S_X,S_Y^+)$ and $s(S_X,S_Y^-)$ are the corresponding matching scores.

All parameters of the model, including the parameters of word embedding, Bi-LSTM, interaction function and MLP, are trained jointly by BackPropagation and Stochastic Gradient Descent. Specifically, we use Adagrad \cite{duchi2011adaptive} on all parameters in training.

\subsection{Discussions}

MV-LSTM can cover LSTM-RNN \cite{palangi2015deep} as a special case. Specifically, if we only consider the last positional sentence representation of each sentence, generated by a single directional LSTM, MV-LSTM directly reduces to LSTM-RNN. Therefore, MV-LSTM is more general and has the ability to leverages more positional sentence representations for matching, as compared with LSTM-RNN.

MV-LSTM has implicitly taken multiple granularity into consideration. By using Bi-LSTM, which has the ability to involve both long and short term dependencies in representing a sentence, MV-LSTM has the potential to capture important n-gram matching patterns. Furthermore, MV-LSTM is flexible to involve important granularity adaptively, compared with CNN based models using fixed window sizes.

\section{Experiments}
In this section, we demonstrate our experiments on two different matching tasks, question answering (QA) and sentence completion (SC).
\subsection{Experimental Settings}
Firstly, we introduce our experimental settings, including baselines, parameter settings, and evaluation metrics.
\subsubsection{Baselines}
The experiments on the two tasks use the same baselines listed as follows.
\begin{itemize}
\item Random Guess: outputs a random ranking list for testing.
\item BM25: is a popular and strong baseline for information retrieval \cite{robertson1995okapi}.
\item ARC-I: uses CNNs to construct sentence representations and relies on a MLP to produce the final matching score \cite{DBLP:conf/nips/HuLLC14}.
\item ARC-II: firstly generates local matching patterns, and then composites them by multiple convolution layers to produce the matching score \cite{DBLP:conf/nips/HuLLC14}.
\item CNTN: is based on the structure of ARC-I, but further uses a tensor layer to compute the matching score, instead of a MLP \cite{DBLP:conf/ijcai/QiuH15}.
\item LSTM-RNN: adopts a LSTM to construct sentence representations and uses cosine similarity to output the matching score \cite{palangi2015deep}.
\item RAE: relies on a recursive autoencoder to learn multiple levels' representations \cite{socher2011dynamic}.
\item DeepMatch: considers multiple granularity from the perspective of topics, obtained via LDA\cite{DBLP:conf/nips/LuL13}.
\item MultiGranCNN: first uses CNNs to obtain word, phrase and sentence level representations, and then computes the matching score based on the interactions among all these representations \cite{DBLP:conf/acl/YinS15}.
\end{itemize}

We can see that ARC-I, CNTN and LSTM-RNN are all single sentence representation models, while ARC-II, DeepMatch, RAE and MultiGranCNN represent a sentence with multiple granularity.

\subsubsection{Parameter settings}
Word embeddings required in our model and some other baseline deep models are all initialized by SkipGram of word2vec \cite{mikolov2013distributed}.
For SC, word embedding are trained on Wiki corpus\footnote{\url{http://nlp.stanford.edu/data/WestburyLab.wikicorp.201004.txt.bz2}} for directly comparing with previous works. For QA, word embeddings are trained on the whole QA dataset. The dimensions are all set to 50. Besides, the  hidden representation dimensions of LSTMs are also set to 50. The batchsize of SGD is set to 128 for both tasks. All other trainable parameters are initialized randomly by uniform distribution with the same scale, which is selected according to the performance on validation set ((-0.1,0.1) for both tasks). The initial learning rates of AdaGrad are also selected by validation (0.03 for QA and 0.3 for SC).

\subsubsection{Evaluation Metrics}
Both tasks are formalized as a ranking problem. Specifically, the output is a ranking list of sentences according to the descending order of matching scores. The goal is to rank the positive one higher than the negative ones. Therefore, we use Precision at 1 (denoted as P@1) and Mean Reciprocal Rank (MRR) as evaluation metrics. Since there is only one positive example in a list, P@1 and MRR can be formalized as follows,
$$ P@1=\frac{1}{N}\sum_{i=1}^{N}\delta(r(S_Y^{+(i)})= 1),$$
$$MRR=\frac{1}{N}\sum_{i=1}^{N}\frac{1}{r({S_{Y}^{+(i)}})},$$
where $N$ is the number of testing ranking lists, $S_Y^{+(i)}$ is the positive sentence in the $i-th$ ranking list, $r(\cdot)$ denotes the rank of a sentence in the ranking list, and $\delta$ is the indicator function.

\subsection{Question Answering}
Question answering (QA) is a typical task for semantic matching. In this paper, we use the dataset\footnote{\url{http://webscope.sandbox.yahoo.com/catalog.php?datatype=l& did=10}} collected from Yahoo! Answers which is a community question answering system where some users propose questions to the system and other users will submit their answers. The user who proposes the question will decide which one is the best answer. The whole dataset contains 142,627 (question, answer) pairs, where each question is accompanied by its best answer.  We select the pairs in which questions and their best answers both have a length between 5 and 50.
After that, we have 60,564 (questions, answer) pairs which form the positive pairs.
Negative sampling is adopted to construct the negative pairs.
Specifically for each question, we first use its best answer as a query to retrieval the top 1,000  results from the whole answer set, with Lucene\footnote{\url{http://lucene.apache.org}}. Then we randomly select 4 answers from them to construct the negative pairs.
At last, we separate the whole dataset to the training, validation and testing data with proportion 8:1:1. Table~\ref{tb:dataset_example} gives an example of the data.

\newcommand{\tabincell}[2]{\begin{tabular}{@{}#1@{}}#2\end{tabular}}
\begin{table}[t]
\centering
\caption{Examples of QA dataset.}
\label{tb:dataset_example}
\resizebox{!}{1.25cm}{
\begin{tabular}{ll} 		   \hline
$S_X$ & \tabincell{l}{\em How to get rid of \textbf{memory stick error} of my sony \\\em  cyber shot?} \\ \hline\hline
$S_Y^+$ & \tabincell{l}{\em You might want to try to format the \textbf{ memory stick} \\ \em but what is the \textbf{error} message you are receiving.}\\ \hline
$S_Y^-$ & \tabincell{l}{\em Never heard of stack underflow \textbf{error}, overflow yes, \\\em overflow is due to running out of virtual \textbf{memory}.}\\ \hline
\end{tabular}}
\end{table}

\begin{table}[t]
\centering
\caption{The effect of pooling parameter $k$ on QA.}
\label{tb:poolingparameter}
\begin{tabular}{lcc} \hline
					& P@1   & MRR 	\\ \hline
LSTM-RNN   		    & 0.690 	& 0.822 \\
Bi-LSTM-RNN       	& 0.702 & 0.830 \\ \hline
MV-LSTM ($k$=1)	& 0.726 & 0.843 \\
MV-LSTM ($k$=3)	& 0.736 & 0.849 \\
MV-LSTM ($k$=5)	& 0.739 & \textbf{0.852} \\
MV-LSTM ($k$=10)& \textbf{0.740} & \textbf{0.852} \\
\hline
\end{tabular}
\end{table}

(1) {\bf Analysis of Different Pooling Parameters}
As introduced in our approach, pooling parameter $k$ is meaningful for our model. If we set $k$ to 1, we can obtain the best matching position for two sentences. While if $k$ is larger than 1, we are leveraging multiple matching positions to determine the final score. Therefore, we conduct experiments to demonstrate the influences of different pooling parameters. Here, the interaction function is fixed to cosine in order to directly compare with the baseline model LSTM-RNN, similar results can be obtained for other interaction functions such as bilinear and tensor layer.

In this experiment, we report different results when $k$ is set to 1, 3, 5 and 10. As shown in Table~\ref{tb:poolingparameter},  the performance is better when larger $k$ is used in $k$-Max pooling for our MV-LSTM. It means that multiple sentence representations do help matching. We also observe that when $k$ is larger than 5, the improvement is quite limited. Therefore, we set $k$ to 5 in the following experiments.

We further compare our model with LSTM-RNN and Bi-LSTM-RNN, where they use LSTM from one and two directions to generate sentence representations, respectively. That is to say, LSTM-RNN views the matching problems as matching at the last position, while Bi-LSTM-RNN both leverage matchings at the first and last position. From the results in Table~\ref{tb:poolingparameter}, we can see that all our MV-LSTMs can beat them consistently. The results indicate that the best matching position is not always in the first or last one. Therefore, the consideration of multiple positional sentence representations is necessary.

We further conduct a case study to show some detailed analysis. Considering the positive pair $(S_X,S_Y^+)$ in Table~\ref{tb:dataset_example}, if $k$ is set to 1, the interaction our model pools out is happened at position $6$ and $9$ in $S_X$ and $S_Y^+$, respectively. The corresponding words at the positions are {\em ``memory"} and {\em ``memory"}. It means that the matching of these two sentences is best modeled when attending to these two words. Clearly the best matching position in this case is not the last one, as implicitly assumed in LSTM-RNN. If $k=5$, the matching positions\footnote{Here, we use the corresponding word at the position to indicate a position.} are (``{\em memory}", ``{\em memory}", 0.84), (``{\em error}", ``{\em error}", 0.81), (``{\em stick}", ``{\em stick}", 0.76), (``{\em stick}", ``{\em memory}", 0.65),  (``{\em memory}", ``{\em stick}", 0.63), with the number stands for the interaction produced by the similarity function. We can see that our model focuses on the keyword correctly and the matching is largely influenced by the positional representations on these keywords. In addition, we also observe that the interactions between {\em ``stick"} and {\em ``memory"} play an important role for the final matching. Therefore, our model can capture important n-gram matching patterns by involving rich context to represent local information.
\begin{table}[t]
\centering
\caption{Experimental results on QA.}
\label{tb:qa}
\begin{tabular}{lcc} \hline
\multicolumn{1}{c}{Model} &
\multicolumn{1}{c}{P@1} &
\multicolumn{1}{c}{MRR} \\ \hline
Random Guess			& 0.200  & 0.457    \\
BM25        			& 0.579 & 0.726 	    \\ \hline
ARC-I 					& 0.581 & 0.756 		\\
CNTN					& 0.626 & 0.781 		\\
LSTM-RNN   		  		& 0.690 	& 0.822 		\\ \hline
RAE	 					& 0.398 & 0.652		\\
DeepMatch  				& 0.452 & 0.679		\\
ARC-II 					& 0.591 & 0.765 		\\
MultiGranCNN				& 0.725 	& 0.840 		\\ \hline
MV-LSTM-Cosine & 0.739 & 0.852 \\
MV-LSTM-Bilinear & 0.751 & 0.860 \\
MV-LSTM-Tensor  		& \textbf{0.766} 	& \textbf{0.869}		\\
\hline
\end{tabular}
\end{table}

(2) {\bf Performance Comparison}
We compare our model with all other baselines on the task of QA. Since there are three different interaction functions, our model has three versions, denoted as MV-LSTM-Cosine, MV-LSTM-Bilinear and MV-LSTM-Tensor, respectively. The experimental results are listed in Table~\ref{tb:qa}.
From the results, we have several experimental findings. Firstly, all end to end deep models (i.e., all baselines except for RAE and DeepMatch) outperform BM25. This is mainly because deep models can learn better representations and deal with the mismatch problem effectively.
Secondly, comparing our model with single sentence representation deep models, such as ARC-I, CNTN, and LSTM-RNN, we can see that all our three models are better than them. Specifically, MV-LSTM-Tensor obtains 11.1\% relative improvement over LSTM-RNN on P@1.
This is mainly because our multiple positional sentence representations can capture more detailed local information than them. Thirdly, comparing our model with multiple granularity models such as RAE, DeepMatch, ARC-II, and MultiGranCNN, our model also outperforms them. Specifically, MV-LSTM-Tensor obtains 5.6\% relative improvement over MultiGranCNN on P@1.
The reason lies in that our local information are obtained by representing the whole sentence, therefore, rich context information can be leveraged to determine the importance of different local information. Finally, among our three models, MV-LSTM-Tensor performs best. This is mainly because tensor layer can capture more complicated interactions, which is consistent with the observation that CNTN outperforms ARC-I significantly.
\begin{table}[t]
\centering
\caption{Case study on QA to compare MV-LSTM with MultiGranCNN.}
\label{tb:casestudy}
\resizebox{!}{1.62cm}{
\begin{tabular}{ll} 		   \hline
$S_X$ & \tabincell{l}{\em How could learn Russian by Internet for \textbf{free}?\\ \em Any good websites?} \\ \hline\hline
$S_Y^+$ & \tabincell{l}{\em Not sure \textbf{free} will get you unforgettable languages, \\ \em however that will give you some basic vocabulary \\\em and great system for remembering it.}\\ \hline
$S_Y^-$ & \tabincell{l}{\em In the Yahoo! home page you will get whole list of \\ \em sites offering this game for \textbf{free} or visit \\ \em www.iwin.com  for \textbf{free} download.}\\ \hline
\end{tabular}}
\end{table}

We give an example in the data, as illustrated in Table~\ref{tb:casestudy}, to further show why our model outperforms the best model which considers multiple granularity, i.e.~MultiGranCNN.
Our experiments show that MultiGranCNN is largely influenced by the word level matching ``{\em free}" to ``{\em free}", and thus get a wrong answer. This is because the two ``{\em free}"s are of different meanings, the first one is focusing on free language resources, while the second one is talking about free games. Therefore, the word/phrase level matching requires to consider the whole context. Our model can tackle this problem by considering multiple positional sentence representations. Specifically, the positional interactions ``({\em free}, {\em free})" is large in matching $S_X$ and $S_Y^+$, while it is small in matching $S_X$ and $S_Y^-$, , which is consistent with our intuitive understanding for matching.

\subsection{Sentence Completion}
In this section, we show our experimental results on sentence completion, which tries to match the first and the second clauses in the same sentence.
We are using exactly the same dataset constructed in \cite{DBLP:conf/nips/HuLLC14} from Reuters \cite{Lewis:2004:RNB:1005332.1005345}.
Specifically, the sentences which have two ``balanced" clauses (with 8-28 words) divided by a comma are extracted from the original Reuters dataset and the two clauses form a positive matching pair. For negative examples, the first clause are kept and the second clauses are sampled from other clauses which are similar with it by cosine similarity.
For each positive example, 4 negative examples are constructed, and thus we will also get 20\% for P@1 by random guess.

The experimental results are listed in Table~\ref{tb:sc}. Considering we are using the same data, some baseline results are directly cited from \cite{DBLP:conf/nips/HuLLC14}, such as ACR-I, ARC-II, RAE and DeepMatch. Since Hu et al. only used P@1 for evaluation in their paper, the results of MRR for these baselines are missing in Table~\ref{tb:sc}.
From the results, we can see that deep models gain larger improvements over BM25, compared with that on QA. This is because the mismatch problem is more serious on this dataset, and usually the first clause has few same keyword with the second one. The other results are mainly consistent with those on QA, and MV-LSTM still performs better than all baseline methods significantly, with 11.4\% relative improvement on P@1 over the strongest baseline.
\begin{table}[t]
\centering
\caption{Experimental results on SC.}
\label{tb:sc}
\begin{tabular}{lcc} \hline
\multicolumn{1}{c}{Model}   & P@1 & MRR \\ \hline
Random Guess 				& 0.200 & 0.457 \\
BM25					 	& 0.346 & 0.568 \\ \hline
ARC-I (Hu et al., 2014)		& 0.475 & - \\
CNTN						& 0.525 & 0.722 \\
LSTM-RNN 					& 0.608 & 0.772 \\ \hline
RAE	(Hu et al., 2014)		& 0.258 & - \\
DeepMatch (Hu et al., 2014)	& 0.325 & - \\
ARC-II (Hu et al., 2014)	    & 0.496 & - \\
MultiGranCNN 					& 0.595 & 0.763 \\ \hline
MV-LSTM-Cosine					& 0.665 & 0.808 \\
MV-LSTM-Bilinear			& 0.679 & 0.817 \\
MV-LSTM-Tensor 				& \textbf{0.691} & \textbf{0.824} \\
\hline
\end{tabular}
\end{table}

\section{Conclusions}
In this paper, we propose a novel deep architecture for matching two sentences with multiple positional sentence representations, namely MV-LSTM. One advantage of our model lies in that it can both capture the local information and leverage rich context information to determine the importance of local keywords from the whole sentence view. Our experimental results and case studies show some valuable insights: (1) Under the assumption that the final matching is solely determined by one interaction (i.e.~pooling parameter is fixed to 1), MV-LSTM can achieve better results than all single sentence representation methods including LSTM-RNN. This means that the best matching position does not always lie in the last one, therefore, the consideration of multiple positional sentence representations is necessary. (2) If we allow the aggregation of multiple interactions (i.e.~pooling parameter is fixed to larger than 1), MV-LSTM can achieve even better results. This means that the matching degree is usually determined by the combination of matchings at different positions. Therefore, it is much more effective by considering multiple sentence representations. (3) Our model is also better than multiple granularity methods, such as DeepMatch, RAE and MultiGranCNN. This means that the consideration of multi-granularity need to rely on rich context of the whole sentence.

\newpage
\section{Acknowledgments}
This work was funded by 973 Program of China under Grants No.~2014CB340401 and 2012CB316303, 863 Program of China under Grant No.~2014AA015204, the National Natural Science Foundation of China (NSFC) under Grants No.~61472401, 61433014, 61425016, 61203298, and 61425016, Key Research Program of the Chinese Academy of Sciences under Grant No.~KGZD-EW-T03-2, and Youth Innovation Promotion Association CAS under Grant No.~20144310.
We also would like to thank Prof. Chengxiang Zhai for the constructive comments.  
\bibliographystyle{aaai}
\bibliography{aaai16_edit}

\end{document}